\documentclass[10pt,twocolumn,letterpaper]{article}

\usepackage{cvpr}
\usepackage{times}
\usepackage{epsfig}
\usepackage{graphicx}
\usepackage{amsmath}
\usepackage{amssymb}
\usepackage{tabularx}
\usepackage{booktabs}

\usepackage[breaklinks=true,bookmarks=false]{hyperref}

\cvprfinalcopy 

\def\cvprPaperID{39} 

\ifcvprfinal\fi
\begin{document}

\newcommand*{\affaddr}[1]{#1} 
\newcommand*{\affmark}[1][*]{\textsuperscript{#1}}
\newcommand*{\email}[1]{\texttt{#1}}

\title{The 3rd Person in Context (PIC) Workshop and Challenge at CVPR 2021:\\
Short-video Face Parsing Track\\
\textnormal{Technical Report:}\\
\textnormal{Shuffle Transformer with Feature Alignment for Video Face Parsing}}

\author{%
    Rui Zhang, Yang Han, Zilong Huang, Pei Cheng, Guozhong Luo, Gang Yu, Bin Fu \\
    \affaddr{Tencent GY-Lab}\\
    \email{\tt\small\{rainarzhang, alanyhan, zilonghuang, peicheng, alexantaluo, skicyyu, brianfu\}@tencent.com}
}

\maketitle

\begin{abstract}
    This is a short technical report introducing the solution of the Team TCParser for Short-video Face Parsing Track of The 3rd Person in Context (PIC) Workshop and Challenge at CVPR 2021.
    
    In this paper, we introduce a strong backbone which is cross-window based Shuffle Transformer for presenting accurate face parsing representation. To further obtain the finer segmentation results, especially on the edges, we introduce a Feature Alignment Aggregation (FAA) module. It can effectively relieve the feature misalignment issue caused by multi-resolution feature aggregation. Benefiting from the stronger backbone and better feature aggregation, the proposed method achieves 86.9519$\%$ score in the Short-video Face Parsing track of the 3rd Person in Context (PIC) Workshop and Challenge, ranked \textbf{the first place}.

\end{abstract}

\section{Introduction}
Face parsing has been applied in a variety of scenarios such as face understanding, editing, synthesis, and animation. As a particular task in semantic segmentation, Face parsing assigns different labels to the corresponding regions on human faces, e.g., hair, facial skins, eyes, nose, mouth and etc., which relies highly on the accuracy of facial components' representation. As a consequence, this is very important to find a way to improve the representation ability. 

Traditionally, ~\cite{smith2013exemplar,warrell2009labelfaces,kae2013augmenting} use hand crafted features including SIFT or machine learning method including Restricted Boltzmann Machine (RBM) to extract local and global features. Current state-of-the-art semantic segmentation approaches ~\cite{chen2017deeplab,yu2018learning, zhao2017pyramid,yu2015multi,peng2017large,huang2019ccnet,ruan2019devil,li2020self,cheng2019spgnet, cheng2020panoptic} based on the fully convolutional network (FCN)~\cite{long2015fully} have made remarkable progress.
For face parsing, CNN-based features are imported into network to extract multi-scale or independent-part facial features~\cite{liu2015multi,luo2012hierarchical,zhou2015inter}. Lin et al. propose two branches with the local-based for inner facial components and the global based for outer facial ones~\cite{lin2019face}. Te et al. propose to learn graph representations over facial images, which model the relations between regions~\cite{te2020edge}. These local-based methods almost adopt the coarse-to-fine strategy, training separated models for various facial components (e.g. eyes, nose etc.) to extract features for each part individually. However, this kind of method achieves good performance at the expense of large memory and computation consumption.

In this paper, we introduce a cross-window based Shuffle Transformer with feature alignment for better improving the face parsing representation ability. Transformers~\cite{dosovitskiy2020image,nicolas2020end,hugo2020training,liu2021swin} have achieve excellent performance on a wide range of visual tasks including image-level classification, object detection, and semantic segmentation. And Shuffle Transformer~\cite{huang2021shuffle} is confirmed as a strong backbone for many vision tasks. Furthermore we import a feature alignment aggregation module to pass information between high-resolution and low-resolution feature maps precisely. The proposed method achieves a good performance in the Short-video Face Parsing track (SFP) of the 3rd Person in Context (PIC) Workshop and achieved the accuracy of 86.9519$\%$, ranked the first place.

\begin{figure*}[htp]
    \centering
    \includegraphics[width=1.0\linewidth]{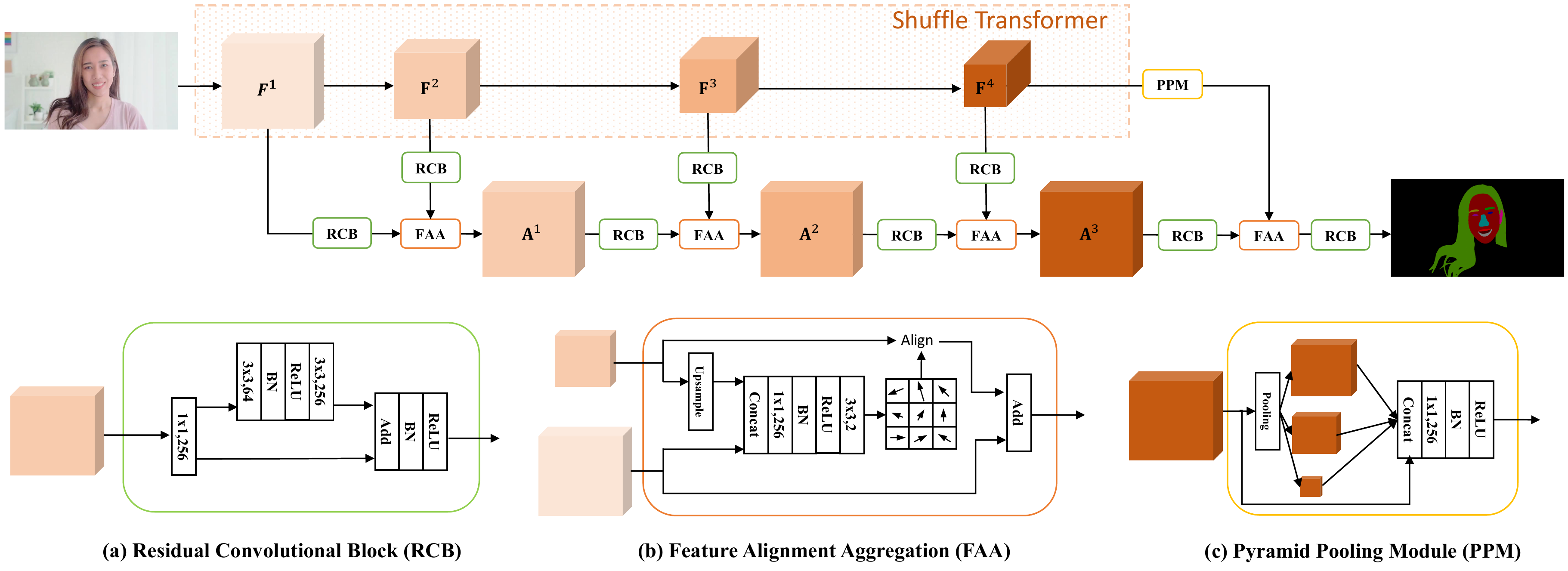}
    \caption{The proposed network architecture.}
    \label{fig:galaxy}
    \vspace{-2mm}
\end{figure*}

\section{Method}
The overall network architecture is visualized in Figure~\ref{fig:galaxy}, which is similar with the AlignSeg~\cite{huang2021alignseg}. Different from ~\cite{huang2021alignseg}, we use Shuffle Transformer~\cite{huang2021shuffle} as a strong backbone and simplified Aligned Feature Aggregation for aligning features from different stages with different resolutions. The simplified Aligned Feature Aggregation, named as Feature Alignment Aggregation (FAA), only learns the offsets maps $\Delta$ for aligning low-resolution high-level feature maps. Meanwhile, we append the Pyramid pooling module (PPM)~\cite{zhao2017pyramid} on the top of the output of the Shuffle Transformer. The $\mathbf{F}^i, i=\{1,2,3,4\}$ are the output feature maps from the $i$th stage of Shuffle Transformer, where $\mathbf{F}^{1} \in \mathbb{R}^{C \times \frac{H}{4} \times \frac{W}{4}}, \mathbf{F}^{2} \in \mathbb{R}^{2C \times \frac{H}{8} \times \frac{W}{8}}, \mathbf{F}^{3} \in \mathbb{R}^{3C \times \frac{H}{16} \times \frac{W}{16}}, \mathbf{F}^{4} \in \mathbb{R}^{4C \times \frac{H}{32} \times \frac{W}{32}}$. The aligned feature $\mathbf{A}^i$ has the same resolution as $\mathbf{F}^{1}$. Finally, the feature is passed through several convolutional layers to produce the final segmentation map.
    
\begin{table}[!htp]
    \renewcommand{\arraystretch}{1.3}
    \setlength{\tabcolsep}{1em}
    \centering
    \caption{Results of SFP, we achieve 86.95\%, win the 1st place.}
    \label{tab:1}
    \begin{tabular}{lcccc}
    \toprule[0.1em]
    Team  & Ranking  & Accuracy(\%) \\
    \toprule[0.1em]
    TCParser(ours)  & 1st & 86.95 \\
    BUPT-CASIA  & 2nd & 86.84 \\
    rat  & 3rd & 86.16 \\
    \bottomrule
    \end{tabular}
    \vspace{-1mm}
\end{table}

\vspace{-0.5cm}

\section{Experiments}
\subsection{Implementation details}
\begin{flushleft}
\textbf{Datasets.}
The total dataset of the SFP challenge contains 1500 videos, each video has 20 images (1 frame per second for each video). SFP has 19535 images for training, 2653 images for validation and 2525 images for testing with 18 categories including face, eye, noes, etc.\\
\vspace{3pt}
\textbf{Evaluation Metrics.}
In this challenge, each submission is validated based on the Davis J /F score and temporal decay.\\
\vspace{3pt}
\textbf{Training.}
We train the proposed model on the open-source machine learning library Pytorch. We use SGD optimizer with 0.9 momentum and 5e-4 weight decay, learning is scheduled via cosine warm up from 7e-3 warmup learning rate at 10 epochs and start cycle at 100 epochs, stop at 150 epochs. Model input size is 672x672, with 6 batch size on eight NVIDIA V100 GPUs. For backbone we use shuffle transformer(pretrained on ImageNet), and the decoder part is trained from scratch. Besides shuffle transformer, we also train different backbone model based on HRNet-W48 for model ensemble and face detection since most of the faces in SFP dataset only account for a small part of the whole image.\\
\vspace{3pt}
\textbf{Testing.}
We add val set into training process, use HRNet-W48 as face dectector, apply multi-scale test, flip, and model ensemble to improve results, as shown in table~\ref{tab:2}.\\
\end{flushleft}

\begin{table}[!htp]
    \renewcommand{\arraystretch}{1.3}
    \setlength{\tabcolsep}{0.1em}
    \centering
    \caption{The effect of our models on val set.}
    \label{tab:2}
    \begin{tabular}{lcccc}
    \toprule[0.1em]
       & mIoU & J\&F-Mean & J-Decay & F-Decay\\
    \toprule[0.1em]
    HRNet-W48 & 0.7283 & 0.900 & 0.004 & 0.005 \\
    Shuffle Transformer & 0.7312 & 0.903 & 0.005 & 0.006 \\
    Model ensemble & 0.7338 & 0.925 & 0.004 & 0.004 \\
    \bottomrule
    \end{tabular}
    \vspace{-0mm}
\end{table}
\vspace{-0.5cm}
\subsection{Experimental Results}
As shown in Table~\ref{tab:1}, our proposed method achieves Accuray score of 86.95\% on the challenge test set and ranks first place. Also we evaluate the J/F score and temporal decay of the proposed models as shown in table~\ref{tab:2}\\
\vspace{-0.5cm}
\section{Conclusion}
In this paper, we use shuffle transformer as backbone and embed FAA module, which learns the offset maps for aligning low-resolution high-level feature maps. With other tricks mentioned, the proposed method achieves 86.95\% accuracy and wins the 1st place in the Short-video Face Parsing track of the 3rd Person in Context (PIC) Workshop. This work hopes to explore a strong feature extractor as well as refine module for face parsing task, we believe it could equally well apply to other similar semantic segmentation tasks.\\

{\small
\bibliographystyle{ieee_fullname}
\bibliography{reference}

\begin{thebibliography}{10}\itemsep=-1pt

\bibitem{nicolas2020end}
Nicolas Carion, Francisco Massa, Gabriel Synnaeve, Nicolas Usunier, Alexander
  Kirillov, and Sergey Zagoruyko.
\newblock End-to-end object detection with transformers.
\newblock In {\em European Conference on Computer Vision}, pages 213--229.
  Springer, 2020.

\bibitem{chen2017deeplab}
Liang-Chieh Chen, George Papandreou, Iasonas Kokkinos, Kevin Murphy, and Alan~L
  Yuille.
\newblock Deeplab: Semantic image segmentation with deep convolutional nets,
  atrous convolution, and fully connected crfs.
\newblock {\em IEEE transactions on pattern analysis and machine intelligence},
  40(4):834--848, 2017.

\bibitem{cheng2019spgnet}
Bowen Cheng, Liang-Chieh Chen, Yunchao Wei, Yukun Zhu, Zilong Huang, Jinjun
  Xiong, Thomas~S Huang, Wen-Mei Hwu, and Honghui Shi.
\newblock Spgnet: Semantic prediction guidance for scene parsing.
\newblock In {\em Proceedings of the IEEE/CVF International Conference on
  Computer Vision}, pages 5218--5228, 2019.

\bibitem{cheng2020panoptic}
Bowen Cheng, Maxwell~D Collins, Yukun Zhu, Ting Liu, Thomas~S Huang, Hartwig
  Adam, and Liang-Chieh Chen.
\newblock Panoptic-deeplab: A simple, strong, and fast baseline for bottom-up
  panoptic segmentation.
\newblock In {\em Proceedings of the IEEE/CVF Conference on Computer Vision and
  Pattern Recognition}, pages 12475--12485, 2020.

\bibitem{dosovitskiy2020image}
Alexey Dosovitskiy, Lucas Beyer, Alexander Kolesnikov, Dirk Weissenborn,
  Xiaohua Zhai, Thomas Unterthiner, Mostafa Dehghani, Matthias Minderer, Georg
  Heigold, Sylvain Gelly, et~al.
\newblock An image is worth 16x16 words: Transformers for image recognition at
  scale.
\newblock {\em arXiv preprint arXiv:2010.11929}, 2020.

\bibitem{huang2021shuffle}
Zilong Huang, Youcheng Ben, Guozhong Luo, Pei Cheng, Gang Yu, and Bin Fu.
\newblock Shuffle transformer: Rethinking spatial shuffle for vision
  transformer.
\newblock {\em arXiv preprint arXiv:2106.03650}, 2021.

\bibitem{huang2019ccnet}
Zilong Huang, Xinggang Wang, Lichao Huang, Chang Huang, Yunchao Wei, and Wenyu
  Liu.
\newblock Ccnet: Criss-cross attention for semantic segmentation.
\newblock In {\em Proceedings of the IEEE/CVF International Conference on
  Computer Vision}, pages 603--612, 2019.

\bibitem{huang2021alignseg}
Zilong Huang, Yunchao Wei, Xinggang Wang, Humphrey Shi, Wenyu Liu, and Thomas~S
  Huang.
\newblock Alignseg: Feature-aligned segmentation networks.
\newblock {\em IEEE Transactions on Pattern Analysis and Machine Intelligence},
  2021.

\bibitem{kae2013augmenting}
Andrew Kae, Kihyuk Sohn, Honglak Lee, and Erik Learned-Miller.
\newblock Augmenting crfs with boltzmann machine shape priors for image
  labeling.
\newblock In {\em Proceedings of the IEEE conference on computer vision and
  pattern recognition}, pages 2019--2026, 2013.

\bibitem{li2020self}
Peike Li, Yunqiu Xu, Yunchao Wei, and Yi Yang.
\newblock Self-correction for human parsing.
\newblock {\em IEEE Transactions on Pattern Analysis and Machine Intelligence},
  2020.

\bibitem{lin2019face}
Jinpeng Lin, Hao Yang, Dong Chen, Ming Zeng, Fang Wen, and Lu Yuan.
\newblock Face parsing with roi tanh-warping.
\newblock In {\em Proceedings of the IEEE/CVF Conference on Computer Vision and
  Pattern Recognition}, pages 5654--5663, 2019.

\bibitem{liu2015multi}
Sifei Liu, Jimei Yang, Chang Huang, and Ming-Hsuan Yang.
\newblock Multi-objective convolutional learning for face labeling.
\newblock In {\em Proceedings of the IEEE Conference on Computer Vision and
  Pattern Recognition}, pages 3451--3459, 2015.

\bibitem{liu2021swin}
Ze Liu, Yutong Lin, Yue Cao, Han Hu, Yixuan Wei, Zheng Zhang, Stephen Lin, and
  Baining Guo.
\newblock Swin transformer: Hierarchical vision transformer using shifted
  windows.
\newblock {\em arXiv preprint arXiv:2103.14030}, 2021.

\bibitem{long2015fully}
Jonathan Long, Evan Shelhamer, and Trevor Darrell.
\newblock Fully convolutional networks for semantic segmentation.
\newblock In {\em Proceedings of the IEEE conference on computer vision and
  pattern recognition}, pages 3431--3440, 2015.

\bibitem{luo2012hierarchical}
Ping Luo, Xiaogang Wang, and Xiaoou Tang.
\newblock Hierarchical face parsing via deep learning.
\newblock In {\em 2012 IEEE Conference on Computer Vision and Pattern
  Recognition}, pages 2480--2487. IEEE, 2012.

\bibitem{peng2017large}
Chao Peng, Xiangyu Zhang, Gang Yu, Guiming Luo, and Jian Sun.
\newblock Large kernel matters--improve semantic segmentation by global
  convolutional network.
\newblock In {\em Proceedings of the IEEE conference on computer vision and
  pattern recognition}, pages 4353--4361, 2017.

\bibitem{ruan2019devil}
Tao Ruan, Ting Liu, Zilong Huang, Yunchao Wei, Shikui Wei, and Yao Zhao.
\newblock Devil in the details: Towards accurate single and multiple human
  parsing.
\newblock In {\em Proceedings of the AAAI Conference on Artificial
  Intelligence}, volume~33, pages 4814--4821, 2019.

\bibitem{smith2013exemplar}
Brandon~M Smith, Li Zhang, Jonathan Brandt, Zhe Lin, and Jianchao Yang.
\newblock Exemplar-based face parsing.
\newblock In {\em Proceedings of the IEEE conference on computer vision and
  pattern recognition}, pages 3484--3491, 2013.

\bibitem{te2020edge}
Gusi Te, Yinglu Liu, Wei Hu, Hailin Shi, and Tao Mei.
\newblock Edge-aware graph representation learning and reasoning for face
  parsing.
\newblock In {\em European Conference on Computer Vision}, pages 258--274.
  Springer, 2020.

\bibitem{hugo2020training}
Hugo Touvron, Matthieu Cord, Matthijs Douze, Francisco Massa, Alexandre
  Sablayrolles, and Herv{\'e} J{\'e}gou.
\newblock Training data-efficient image transformers \& distillation through
  attention.
\newblock {\em arXiv preprint arXiv:2012.12877}, 2020.

\bibitem{warrell2009labelfaces}
Jonathan Warrell and Simon~JD Prince.
\newblock Labelfaces: Parsing facial features by multiclass labeling with an
  epitome prior.
\newblock In {\em 2009 16th IEEE international conference on image processing
  (ICIP)}, pages 2481--2484. IEEE, 2009.

\bibitem{yu2018learning}
Changqian Yu, Jingbo Wang, Chao Peng, Changxin Gao, Gang Yu, and Nong Sang.
\newblock Learning a discriminative feature network for semantic segmentation.
\newblock In {\em Proceedings of the IEEE conference on computer vision and
  pattern recognition}, pages 1857--1866, 2018.

\bibitem{yu2015multi}
Fisher Yu and Vladlen Koltun.
\newblock Multi-scale context aggregation by dilated convolutions.
\newblock {\em arXiv preprint arXiv:1511.07122}, 2015.

\bibitem{zhao2017pyramid}
Hengshuang Zhao, Jianping Shi, Xiaojuan Qi, Xiaogang Wang, and Jiaya Jia.
\newblock Pyramid scene parsing network.
\newblock In {\em Proceedings of the IEEE conference on computer vision and
  pattern recognition}, pages 2881--2890, 2017.

\bibitem{zhou2015inter}
Yisu Zhou, Xiaolin Hu, and Bo Zhang.
\newblock Interlinked convolutional neural networks for face parsing.
\newblock In {\em International symposium on neural networks}, pages 222--231.
  Springer, 2015.

\end{thebibliography}
}

\end{document}